\documentclass[conference]{IEEEtran}
\IEEEoverridecommandlockouts
\usepackage{amsmath, amssymb, amsfonts,footmisc}
\usepackage{graphicx,xcolor,fdsymbol,fancyhdr}
\usepackage[dvipsnames]{xcolor}
\usepackage{booktabs, rotating,multirow,lscape}
\usepackage{hyperref,float,algorithm2e,tikz}
\usepackage[center]{caption}
\usepackage{cite}
\usetikzlibrary{positioning}
\newcommand{\club}[0]{\clubsuit} 
\renewcommand{\diamond}{\color{red}\vardiamondsuit\color{black}} 
\newcommand{\spade}[0]{\spadesuit} 
\newcommand{\heart}[0]{\color{red}\varheartsuit\color{black}} 

\begin{document}

\title{Quantitative Rule-Based Strategy modeling in Classic Indian Rummy: A Metric Optimization Approach}


\author{\IEEEauthorblockN{Purushottam Saha}
\IEEEauthorblockA{
Indian Statistical Institute, Kolkata}
\and
\IEEEauthorblockN{Avirup Chakraborty}
\IEEEauthorblockA{
Indian Statistical Institute, Kolkata}
\and
\IEEEauthorblockN{Sourish Sarkar}
\IEEEauthorblockA{
Indian Statistical Institute, Bangalore}
\and
\IEEEauthorblockN{Subhamoy Maitra}
\IEEEauthorblockA{
Applied Statistics Unit,\\
Indian Statistical Institute, Kolkata}
\and
\IEEEauthorblockN{Diganta Mukherjee}
\IEEEauthorblockA{
Sampling and Official Statistics Unit,\\
Indian Statistical Institute, Kolkata}
\and
\IEEEauthorblockN{Tridib Mukherjee}
\IEEEauthorblockA{
Chief Data Scientist \\and AI Officer, IDfy}
}


\maketitle

\begin{abstract}
\boldmath
The 13-card variant of Classic Indian Rummy is a sequential game of incomplete information that requires probabilistic reasoning and combinatorial decision-making. This paper proposes a rule-based framework for strategic play, driven by a new hand-evaluation metric termed \emph{MinDist}. The metric modifies the \emph{MinScore} metric by quantifying the edit distance between a hand and the nearest valid configuration, thereby capturing structural proximity to completion. We design a computationally efficient algorithm derived from the \emph{MinScore} algorithm, leveraging dynamic pruning and pattern caching to exactly calculate this metric during play. Opponent hand-modeling is also incorporated within a two-player zero-sum simulation framework, and the resulting strategies are evaluated using statistical hypothesis testing. Empirical results show significant improvement in win rates for \emph{MinDist}-based agents over traditional heuristics, providing a formal and interpretable step toward algorithmic Rummy strategy design.
\end{abstract}

\begin{IEEEkeywords}
Game theory, Rummy, heuristic optimization, opponent modeling, rule-based strategies, zero-sum games.
\end{IEEEkeywords}

\section{Introduction} \label{sec:rs_intro}
Classic Indian Rummy (13 cards) is one of the most widely played card games in India, characterized by a balance between luck and skill. Players aim to form valid melds: sets or sequences of cards, by drawing and discarding over successive turns. The game's decision complexity arises from hidden information, stochastic draws, and the combinatorial explosion of possible card arrangements.

While reinforcement learning (RL) and Monte Carlo methods have been extensively used for other imperfect-information games, Rummy’s structure invites interpretable, rule-based approaches grounded in explicit metrics of hand quality. In this paper, we develop such a rule-based framework that introduces a new quantitative metric, \emph{MinDist}, to guide play.

\emph{MinDist} modifies the existing \emph{MinScore} metric by incorporating combinatorial proximity: the minimum number of card changes required for a hand to become valid. We combine this with a computational trick that exploits bit-masks and the use of super-jokers, making the solution tractable. Further, for the strategy development, we embed heuristic opponent hand modeling, adding empirical domain knowledge/ playing expertise to the strategies for performance improvement. The agents are evaluated within a two-player zero-sum simulation, with hypothesis testing confirming significant improvements in strategic performance than baseline random and \emph{MinScore} based strategies.

The remainder of the paper is organized as follows. Section~II introduces the rules of the game of Rummy. Section~III reviews related work in this domain. Section~IV formalizes the hand evaluation metrics used in this study, while Section~V presents the algorithms developed to compute these metrics. Section~VI describes the heuristic framework for opponent hand modeling. Section~VII outlines the simulation setup and experimental design. Section~VIII reports and analyzes the empirical results, and Section~IX concludes the paper with a discussion of key findings and implications.

\section{Game Rules} \label{sec:rs_rules}
The game of Rummy have had many variations, but there is a very limited literature regarding the widely popular 13 card Classic Indian Rummy. Each player is dealt 13 cards initially; if the number of players is 2, then a 52 cards deck is chosen for the game and if there are 6 players, two decks of 52 cards each is combined for the game. Each player has to draw and discard cards by turns till one player melds their cards with valid sets that meet the Rummy validation rules.

Each game starts with shuffling and dealing of 13 cards to each of the players, and drawing a card as the wildcard joker for the game to set aside (If say, 9$\diamond$ is drawn as the wildcard joker, then all 9's from different suits are recognized as the wildcard jokers of the game, so 7 wildcard jokers for 2 decks).

The game continues with players taking turn to draw a card and discard a card in order to group the hand as a collection of valid melds (of size atleast 3), which are either:
\begin{itemize}
    \item \textbf{Pure Sequences:} A group of cards in sequential order of ranks,  all with the same suit (e.g.,  \{3$\heart$,  4$\heart$,  5$\heart$,  6$\heart$\}).
    
    \item \textbf{Impure Sequences:} A group of cards in sequential order of ranks where a wildcard or printed joker replaces one or more missing cards (e.g.,  \{3$\heart$,  4$\heart$,  9$\club$,  6$\heart$\},  where 9$\club$ is a wildcard joker).
    
    \item \textbf{Pure Sets:} A group of cards with the same rank but different suits (e.g.,  \{8$\heart$,  8$\spade$,  8$\diamond$\}).
    
    \item \textbf{Impure Sets:} A group of cards with the same rank but different suits (e.g.,  \{8$\heart$,  8$\spade$,  9$\club$\}),  where 9$\club$ is a wildcard joker.
\end{itemize}

The player who declares with all valid sequences and sets wins the game, of which the first must be a Pure Sequence and the next one must be a Pure or Impure Sequence (invalid declaration results in immediate loss with 80 points, maximum achievable). Furthermore, at any time during the game, the player has the option of dropping out to conserve points (e.g.,  20 points if dropped in the first round,  else 40), i.e. to secure not-so-harmful losses. Learning when to drop is a vital skill for a good Rummy player. The present version of the game should be concluded in 100 rounds having win-loss outcome only. Although the winner is declared based on the minimum score, or clarified as deadwood score (scores left from cards after valid melds are set apart, keeping in mind the requirement condition),  ties are broken based on total number of points in suit Diamonds,  Clubs,  Hearts and Spades respectively in the very order. If still the tie is not broken,  then the first such player in the order is declared as the winner. 

Note that the conclusion of the game results with the winner winning points as the difference of scores at the terminal stage. However,  special cases of fold implicates the limited gain of 20 or 40 (in case of fold during first move\footnote{can also occur if opponent declares with a valid declaration before the player gets a turn} or second move onwards, respectively). This gain is henceforth denoted as the \hypertarget{def:gain}{\textbf{Gain}} of the winner,  and signed version of the same is used as a metric of player 1 for later simulations,  treating the game as a \textbf{zero-sum game}.

\section{Related Work} \label{sec:rs_literature}
Games have long provided a foundational testbed for artificial intelligence (AI) research. Since the earliest days of the field, they have offered structured environments for studying reasoning, uncertainty, and decision-making (Greenwald et al. 2020). Early advances were grounded in game-theoretic adversarial search, where methods such as Minimax (Russell and Norvig 2009) and Monte Carlo Tree Search (MCTS) (Chaslot et al. 2006) systematically propagated players’ actions through trees of possible game states. To manage \textbf{large state spaces}, depth-limited searches and \textbf{evaluation functions} were introduced to approximate the value of intermediate positions, initially designed through domain knowledge for games like Chess (Shannon 1950), and later learned automatically in complex or stochastic settings such as Backgammon (Tesauro 1995) and Go (Silver et al. 2016).

These classical approaches, however, assume perfect information, where all players observe the full game state. Card games such as Gin Rummy and Classic Indian Rummy instead involve hidden information, arising from unobserved cards in the deck or opponents’ hands, making optimal play significantly more complex. Research on imperfect-information games has been advanced notably in Poker, where agents such as Loki-2 (Billings et al. 1999) employed probabilistic approximations to evaluate hidden states. Loki-2’s Hand Evaluator estimated the strength of its hand relative to possible opponent holdings, integrating belief-based reasoning into strategic evaluation. 

Despite the limited research available on Indian Rummy, several studies have investigated a similar variant, Gin Rummy, showing meaningful progress in understanding its strategic aspects. Eicholtz et al (2021) \cite{heisen.bot.rummy} showed building rule-based agents, and strategic decision making by different metrics\cite{heur.eval.gin.rummy}, such as myopic meld distance \cite{myopic.melds}(a heuristic and modified version of the proposed \emph{MinDist} metric) and estimation of card fitness for discard\cite{discard.paper} has created a rich literature on this variant.  

Building upon these foundations, our work applies quantitative rule-based strategy modeling to Classic Indian Rummy. We introduce a metric optimization framework that formalizes intuitive player heuristics into quantifiable decision criteria, bridging traditional symbolic strategies with modern optimization-based evaluation methods.

\section{Metric Formulation} \label{sec:rs_metrics_form}
For analysis of the game, we next define a few metrics of a given hand, which will help the agents to take decisions throughout a game. We mainly consider the $MinScore$ and the $MinDist$ metrics of a hand which will in turn, guide play.

\subsection{MinScore}
Given a hand $h$, and a wild-card joker $wcj$ for the game, $MinScore(h, wcj)$ returns the optimal (minimum) achievable score of that hand, by optimally grouping the cards into valid melds of Pure/Impure sequences and sets. Hence, at any given point of time, \emph{MinScore} is the amount of loss if the game is to terminate the very moment -- a high \emph{MinScore} implies higher loss (given the game is to terminate in a few moves), a low \emph{MinScore} implies little loss, with a zero \emph{MinScore} suggesting a valid declaration.

For example, consider the hand $3\club 4\club 5\club 6\club \;\; 9\diamond 10\diamond J\diamond \;\; 4\diamond 4\heart 4\spade \;\; K\spade K\heart 7\spade$, with $3\spade$ as wild card joker. It can be seen that the first two melds form valid pure sequences, hence the requirement is complete, but amongst rest of the cards, 3 cards are forming a valid set, while 3 cards (i.e. $K\spade K\heart 7\spade$) are not, so the \emph{MinScore} for the hand is the sum of value of these cards, i.e. 10+10+7 = 27.\\

This problem of optimal grouping can be recursively solved using the following recursion:
\begin{equation}
    MinScore(h, wcj) = \min_{m \in V_{h,S}} MinScore(h \setminus m,\; wcj), 
\end{equation}
where $V_{h,S}$ is the set of all valid melds possible in hand h with the wild-card joker wcj. Generally the maximum score obtainable in a game of Classic Indian Rummy is 80, hence we consider \emph{MinScore} as minimum of the aforementioned quantity and 80 (This makes the \emph{MinScore} the exact points a player obtain if the declaration is optimal). We have to keep in mind that the set of valid melds gets updated in each step of the minimization keeping in mind all the cards that are already chosen (a card can't be chosen twice and the requirement of specific melds (a pure sequence and a pure or an impure sequence for example, denoted by S).

The base cases for this recursion involve hands not having any valid meld as subset and thus setting the min-score to the same as sum of value of all the cards in the hand (2-10 numbered cards get the number same as the face value,  face-cards including Ace get 10 as value,  printed jokers along with wild-card jokers get 0 as value) also known as the \textbf{dead-wood score}.

We simulated 10000 random hands of size 13 from a single deck with 2 printed jokers, along with a wildcard joker and calculated the \emph{MinScore} of the hands. The empirical CDF for them is shown below.

\begin{figure}[!htb]
    \centering
    \includegraphics[width=0.7\linewidth]{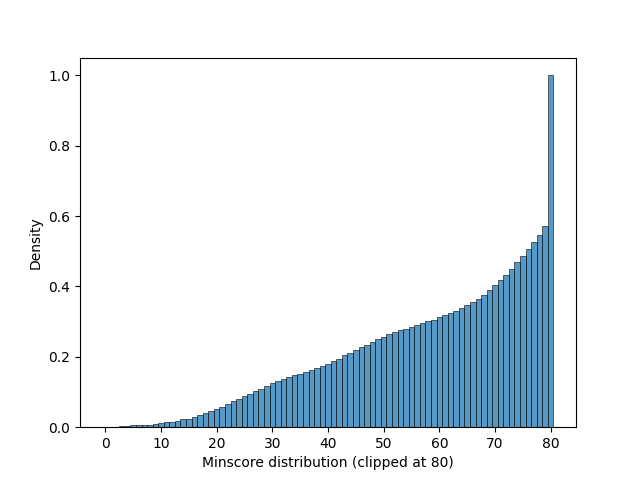}
    \caption{MinScore eCDF (clipped at 80)}
    \label{fig:mscore_eCDF}
\end{figure}

\begin{figure}[!htb]
    \centering
    \includegraphics[width=0.7\linewidth]{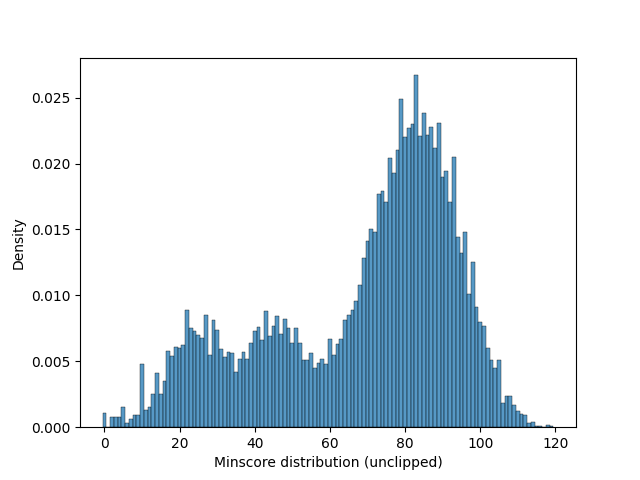}
    \caption{MinScore histogram (unclipped)}
    \label{fig:mscore_unclipped}
\end{figure}

We observe that a significant peak is observed at 80, which also justifies the clipping to occur at 80, and the fact that observed \emph{MinScore} for a randomly selected hand will be exactly or close to 80 with significant probability, this will also be later reflected on. 

As a heuristic, it can be safely assumed that in an optimal strategy, \emph{MinScore} of a hand will improve through each round. This is later to be used as a strategy, introduced as MinScore Agent. The algorithmic challenges of \emph{MinScore} is discussed in the next section.

An important point here to notice is this measure penalizes high-value unmatched cards but fails to reflect proximity to valid configurations (i.e. complete hands). To address this very intuition, we develop another combinatorial metric, \emph{MinDist}. 

\subsection{Proposed MinDist Metric}
We define:
\begin{equation}
MinDist(H) = \min_{H' \in \mathcal{V}} d(H, H'),
\end{equation}
where $\mathcal{V}$ is the set of all valid hands, and $d(H, H')$ is the minimal number of card replacements required to transform $H$ into a valid configuration $H'$. This metric allows discrimination between near-valid hands, enabling better long-term strategy estimation. This structural distance can be understood as minimum number of turns required to be able to reach a valid declaration, which in turn can also help proxy the time to declare for the opponent (along with a suitable scale). \\

We simulated 10000 random hands of size 13 from a single deck with 2 printed jokers, along with a wildcard joker and calculated the \emph{MinDist} of the hands. The histogram is shown below.\\

\begin{figure}[H]
    \centering
    \includegraphics[width=0.7\linewidth]{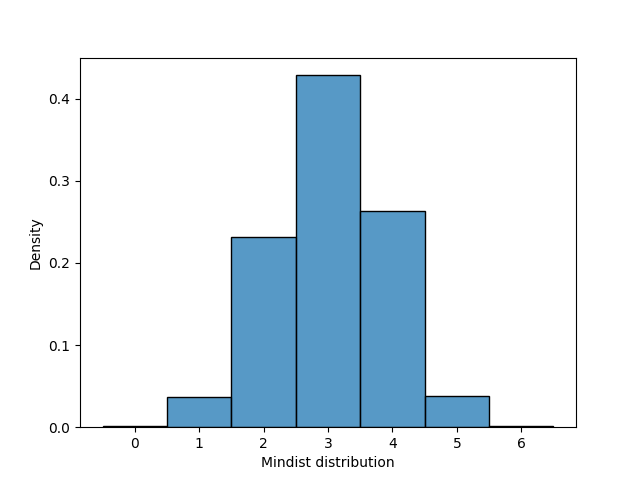}
    \caption{MinDist histogram}
    \label{fig:mindist.hist}
\end{figure}

As observed from the figure~\ref{fig:mindist.hist}, the \emph{MinDist} of a randomly selected hand is between 2 and 4, with very high probability. In the simulation, the highest observed \emph{MinDist} for a hand is 6, where as it can easily be observed that this metric can not exceed 9 (i.e. atmost 9 cards need to be changed to make the hand complete), as any 4 cards can be part of a complete hand, as those 4 cards will guide to create 4 valid melds, each of size 3 with one of size 4, to partition the hand into valid melds. \\ 
The key motivation behind \emph{MinDist} as a metric is that \emph{MinScore}, though it reflects the optimal score achievable if the game ends that very round, it does not have much discriminative power between hands. As the score is capped at 80, a lot of hands have 80 as the MinScore (observed from Figure~\ref{fig:mscore_unclipped}), which makes it difficult to compare between those hands, though one of hands might always be preferable over the other. For example, consider a hand with no pure sequences, i.e. a to-be meld waiting on a card to make it a pure sequence, but all the other cards are partitioned into valid impure sequences. $3\diamond 9\club 5\diamond 6\diamond\;\;J\spade Q\spade 7\club\;\;2\club 2\spade 2\heart\;\;A\club A\spade A\heart$ is such a hand, where $7\club$ is the wild card joker (hence the second grouping is a valid impure sequence). If the $9\club$ card could be replaced with $4\diamond$, then the hand will be a complete hand, and hence the \emph{MinDist} is 1, though the \emph{MinScore} is 80. It is very intuitive that the hand given is favorable to most other hands with 80 \emph{MinScore}. \footnote{considering $4\diamond$ is not in pile, and hence is available for the player to be obtained. This is a serious limitation of \emph{MinDist} i.e. it does not incorporate the fact that cards on the visible pile, are not available. But this is not the focus of the current study, as MinDist also does not consider the same, hence it is still an considerable improvement upon \emph{MinScore}.} Also, in each turn, the \emph{MinDist} can change by atmost 1, making it a more stable metric than \emph{MinScore} of a given hand across a game. \\
Though this observation doesn't nullify the importance of MinScore as a metric, and rather it builds upon that. So in case of two hands with same \emph{MinDist} (which is very common, suppose while obtaining a new card and discarding a card during a turn, the \emph{MinDist} does not change), the \emph{MinScore} might differ, and hence we can prefer one hand than the other. Later we will see applying this idea as a heuristic for a strategy also helps improve the performance of the strategy. 

\section{Algorithmic Framework}

To solve the \emph{MinScore} problem, we target the recursion equation discussed, and try to solve the same via Dynamic Programming. Given a hand, we can calculate the set of valid melds according to given state S (by looping over each 3-5 card tuples and checking for valid melds, as a 6 card tuple which is a valid meld, can be broken into two valid melds of size 3). We define the states into three parts, first one which denotes the first stage: only pure sequences are allowed as valid melds. Given such a pure sequence is found, it transitions into the next state: only impure/pure sequences are allowed. Once this stage is passed, then the next stage allows any possible valid meld that could be formed. This process appears as a branching problem, a hand is branched into subsets of the hand, which is formed from removing cards that make a valid meld as per the state, the terminal nodes have value as sum of scores of each card, and minimum score of children nodes is propagated to the parent node.
\begin{figure}[H]
    \centering
    \includegraphics[width=0.8\linewidth]{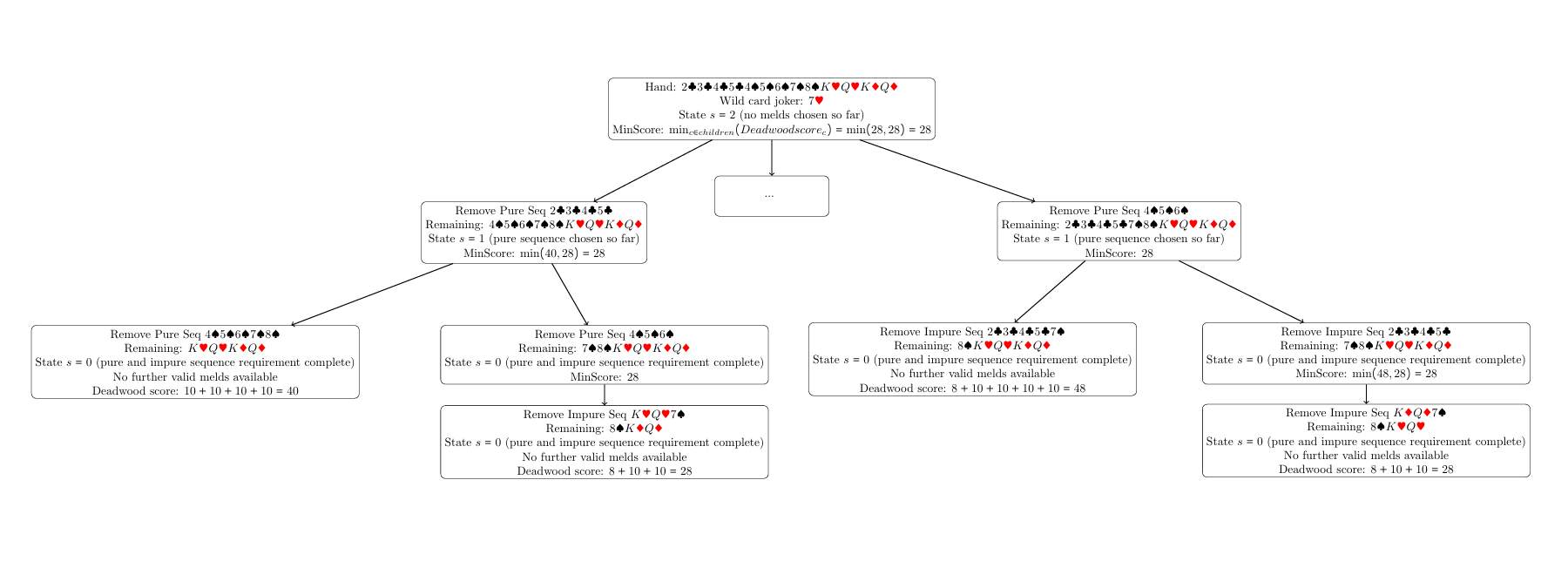}
    \caption{Example MinScore DP transitions for a given hand}
    \label{fig:minscore.example}
\end{figure}
Despite the dynamic programming approach, it becomes extremely expensive in terms of memory and time to calculate valid melds by the primitive functions. To solve this problem, a useful method applied is to \textbf{pre-compute} the total set of valid melds, the only challenge remaining is to restrict that set conditioning on the cards already used for melds in earlier stages, and also based on the state. This is solved by the technique of \textbf{bit-masking}. Bit-masking essentially means storing a subset of the hand (e.g. a valid meld) using 0-1 masking, 1 in $\text{i}^{th}$ position i if $\text{i}^{th}$ card is present n the subset, 0 otherwise, representing any subset of the hand with just 13 bits -- a massive improvement over storing a list of unordered cards. A quick pass can calculate and store masks of all valid melds. Also restricting the set of valid melds to the remaining hand is extremely easy: the \textbf{bitwise AND} operator tells if a meld is still a subset of the remaining hand, and to confirm the meld and further restrict the remaining hand, a \textbf{bitwise XOR} can be applied to make the meld cards unavailable (or 0) in the remaining hand masks. This drastically improves the computation time of a meld - from being intractable to under 100 ms. An implementation of this algorithm can be found in the blog post by Games24x7. \cite{lal2025rummy}

To generalize this algorithm to be able to assist game play and  analyze other variations\footnote{though not elaborated in this paper}, we have introduced a few key parameters. Obtaining an optimal grouping of smaller size removing (say) k cards is very helpful, as it aids the problem of dropping a card after drawing a card as part of a turn. Other generalizations include requirements of the melds (Pure sequence of length 3+ followed by pure or impure sequence of length 3+ for Indian rummy), capping of the maximum obtainable score (80 for 13 card Indian rummy), minimum length for a meld to be valid (3 for our case), declaration of optimal grouping to be toggled on or off (significantly boosts computational time as storing a lot of subset masks is not required to just calculate the \emph{MinScore}) etc. A pseudo code of the algorithm is presented below. A python implementation can be found in the \href{https://github.com/purushottam-saha/Rummy}{github code repository}. 

\begin{algorithm}
\caption{MinScore Algorithm (Simplified)}
\SetAlgoLined
\KwIn{Hand $H$, wild card joker $wcj$, req. melds $req$}
\KwOut{Min possible score and corres. declaration}
\vspace{0.2cm}
\textbf{Initialization:} \\
Compute card values for all cards in $H$\;
Generate all valid melds $M$ from subsets of $H$ 
  satisfying pure/impure seq or set rules (using $wcj$)\;
Initialize dynamic programming (DP) table 
  $\text{dp}[mask][state] \leftarrow \infty$ for all subsets $mask$\;
Set $\text{dp}[0][0] \leftarrow 0$\;

\vspace{0.2cm}
\textbf{Recursive computation:}\\
\For{each subset of cards represented by bitmask $mask$}{
    Compute deadwood score = sum of card values in $mask$\;
    $\text{dp}[mask][state] \leftarrow$ deadwood score\;
    \For{each valid meld $m \in M$}{
        \If{$m$ is a subset of $mask$}{
            $newMask \leftarrow mask \setminus m$\;
            Determine meld type and update requirement state $state'$\;
            $score \leftarrow \text{dp}[newMask][state']$\;
            \If{$score < \text{dp}[mask][state]$}{
                $\text{dp}[mask][state] \leftarrow score$\;
                Store $m$ for backtracking optimal declaration\;
            }
        }
    }
}

\vspace{0.2cm}
\textbf{Backtracking:}\\
Reconstruct optimal meld combination by tracing stored melds 
  from full mask $(1 << |H|) - 1$ back to $0$\;
Compute total score as sum of unmelded card values (deadwood)\;

\vspace{0.2cm}
\textbf{Return:}\\
$(\text{minScore}, \text{optimal declaration})$\;
\end{algorithm}

As \emph{MinScore} is a problem of optimal partition, hence the objective is tractable, while on the other hand, \emph{MinDist} deals with cards from the hand to be replaced by unobserved cards. This seems difficult at the first sight, but with a clever trick (referred as the \textbf{Super Joker trick}) can be reduced to \emph{MinScore} type problem. 

The unobserved cards can be viewed as a separate card that can replace itself with some remaining card from the deck to make valid melds, hence considering itself as a joker is a natural idea, but a joker is unable to form a pure meld -- but as the unobserved card can be any card, it can be replaced to form a pure meld (under some conditions discussed below). This special type cards are called as Super Joker cards: Joker cards which can form pure melds. This can be integrated in the building block boolean functions which check for valid melds, so that the rest algorithm is untouched from that of \emph{MinScore}.

The only standing problem is the fact that the number of these unobserved cards is to be minimized, and throughout the process of \emph{MinScore} calculation it's difficult to keep track of the same. Hence we gradually increase the number of Super Jokers in a hand, keeping all the other cards intact, and try to form a valid declaration (decision version of the \emph{MinScore} problem: \emph{MinScore} = 0 $\iff \exists$ a valid declaration), with 13 cards. If at stage k, i.e. with k many Super Jokers we can obtain a valid declaration, and with j many we can't (where j = 0, 1, $\ldots$, k-1 ), we say the \emph{MinDist} of the hand is k.

\begin{algorithm}
\caption{MinDist Algorithm (Simplified)}
\SetAlgoLined
\KwIn{Current hand $H$, wild card joker $wcj$, req. melds $req$, maximum distance $maxdist$}
\KwOut{Minimum number of cards to add (distance) for a valid declaration, and declaration if applicable}

\textbf{Initialization:} \\
Set minimum sequence length $minlen$ and total requirement levels from $req$\;
\For{all subsets of $H$}{
    Identify all valid melds $M$ (pure/impure sets or sequences using $wcj$)\;
}
Initialize DP table $\text{dp}[mask][state] \leftarrow \infty$ for all subsets of $H$\;
Set $\text{dp}[0][0] \leftarrow 0$\;

\vspace{0.2cm}
\textbf{Core Recursive Function:} \\
Compute(mask, state, cards, needed)\\
    Compute unmatched card count $d = $ number of cards in $mask$\;
    Set $\text{dp}[mask][state] \leftarrow d$\;
    \For{each meld $m \in M$}{
        \If{$m \subseteq mask$}{
            $newMask \leftarrow mask \setminus m$\;
            Determine meld type and update requirement state $state'$\;
            $dist \leftarrow \text{dp}[newMask][state']$\;
            \If{$dist < \text{dp}[mask][state]$}{
                $\text{dp}[mask][state] \leftarrow dist$\;
                Store $m$ for reconstruction\;
            }
        }
    }
    \Return{$\text{dp}[mask][state]$}\;

\vspace{0.2cm}
\textbf{Waste Card Evaluation:}\\
WasteCards(new\_cards)\\
    Recompute valid melds $M$ for $new\_cards$\;
    Initialize DP table as before\;
    $result \leftarrow \text{Compute(fullMask, lastState, new\_cards, needed)}$\;
    \If{declaration required}{
        Backtrack stored melds to form optimal declaration\;
        \Return{$(result, declaration)$}\;
    }
    \Return{$result$}\;

\vspace{0.2cm}
\textbf{Minimum Distance Calculation:}\\
MinDist()\\
    \For{$i = 0$ \KwTo $maxdist$}{
        Create $new\_cards = H + i$ dummy jokers\;
        $r \leftarrow \text{WasteCards}(new\_cards)$\;
        \If{a valid declaration found with total distance $i + shift$}{
            \Return{$(i, declaration)$}\;
        }
    }
\end{algorithm}

As with \emph{MinScore}, similar extensions are acted on \emph{MinDist} as well, such as picking best 13 out of 14 (more straightforward than \emph{MinScore}), different requirement of melds, optimal declaration modulo unobserved cards etc. The aforementioned python implementation in the github repository can be used for the required calculations and possibly game-play assist.


\subsection{On Correctness of the \emph{MinDist} Algorithm}
Suppose adding k many super jokers, we can obtain a collection of valid disjoint melds spanning over 13 cards. Which means { d(H, H') for all H' valid declaration } has k as an element, and hence minimum over the set is $\leq$ k. Now if for all k'< k, we are not able to find a collection of valid disjoint melds spanning over 13 cards, this means $\not\exists \;\text{valid } H' \:$ s.t. $d(H,H') < k$, hence by definition, \emph{MinDist} is exactly k. This proves that the output of the above algorithm is correct. 




\section{Opponent Hand modeling}
While a sophisticated Bayesian hand modeling could have delivered much more promising results, the reason for us to use intuitive hand modeling and based thumb rules is to eradicate the pain of tuning hyper-parameters as the strategies based on \emph{MinDist} were already quite expensive, and we wished to extract as much as possible for elementary strategies. Here for the case of Opponent Hand modeling in 2 player case, we have memorized the cards taken by the Opponent from the open pile and discarded cards to the open pile (public information), in order to infer about the rest of the cards in the hand of the opponent. The inferences were also rule based and very simple, given we have equal preference to drop a card after picking, we would wish not to drop a card for the opponent to take, which can create a meld with the picked up cards by the opponent; and also we would wish to drop a card which creates a meld with the dropped cards by the opponent. Later we would see this simple rule based modeling improved the performance over strategies based on simply \emph{MinDist}.

\section{Simulation Framework}
In this section, we present the simulation framework of empirical analysis of playing strategies against each other. The code for the exercise can be found \href{https://github.com/purushottam-saha/Rummy}{here}.

First we create set of profiles to be studied. These profiles are set of instructions to agents about how to play the game, which are mostly guided by \emph{MinScore} or \emph{MinDist} criteria. The profiles are as follows.

\begin{enumerate}
    \item \textbf{Random}: This agent randomly chooses its decisions from its current set of choices (Choose a card from Deck or Pile, and discard a card randomly; or fold). From our understanding this is a very naive algorithm and hence should have near 0 win rate, except for against itself, or a strategy that seeks to loose. 
    \item \textbf{Defeat seeking (Naive)}: This strategy identifies a meld, if any, from its hand, and attempts to destroy it by throwing the lowest point card,  picking the deck card or pile card (if pile card does not form a meld itself). If a meld does not exist even with pile card,  then it picks the deck card and repeats aligning the goal to maximize the score. This agent acts as a benchmark to exhibit its possible to lose the game if such anti-optimal play is considered: exhibiting that this game has skill component (if such a strategy would not exist then it would gather evidence towards the game having a more prominent luck component than a skill component).
    \item \textbf{\emph{MinScore} based}: This agent considers the pile card, if by optimal grouping (of 13 cards) by \emph{MinScore} algorithm suggests picking of the pile card, then it does so; otherwise it picks the deck card, and removes the card suggested by optimal grouping of best 13 cards guided by \emph{MinScore}. This strategy also has a \textbf{drop adherence} to be turned on,  i.e. it drops its hand if the \emph{MinScore} of the hand is greater than or equal to some value (say 80) in the first round, to cut off big losses.
    
    \item \textbf{\emph{MinDist} based}:This agent considers the pile card, if by optimal grouping (of 13 cards) by \emph{MinDist} algorithm suggests picking of the pile card (\emph{MinDist} of the best 13 cards reduces), then it does so; otherwise it picks the deck card, performs \emph{MinDist} on the new hand and removes the highest score card suggested by optimal grouping of best 13 cards guided by \emph{MinDist} (highest value cards replaced by unobserved cards or Super Jokers). This strategy also has a \textbf{drop adherence} to be turned on,  i.e. it drops its hand if the \emph{MinDist} of the hand is greater than or equal to some value (say 3) in the first round, to cut off big losses.
    
    \item \textbf{\emph{MinDist} improved by \emph{MinScore}}: This agent chooses the pile card if by picking so improves its min-dist,  and if not it similarly acts on the deck card,  but if the min-dist is same,  it removes the card which improves min-score the most. It also has a similar drop adherence as the Min-dist based agent.
    \item \textbf{Min-dist based improved by Opponent play}: This agent is the only one which is loosely based on opponent modeling. This agent plays just as Min-dist based agent,  while it improves the min-dist; but if it does not,  then instead of dropping any card,  it tried to drop the cards (if possible) that are closest to the cards the opponent drops,  and then ranked by value of the cards. Here the term closest to card c means which can help form a valid meld with c. For example,  if the opponent has dropped 5 of Clubs then it will try to first drop 4,  6 of Clubs or 5 of other suits. It also has a similar drop adherence as Min-dist based agent.
\end{enumerate}
The main goal of this simulation is the strategy and skill in game navigating,  and hence all players are assumed to declare their hand when they have a declarable hand,  and all the analysis must be done with this understanding. Also because of the same reasoning,  score and \emph{MinScore} is used interchangeably throughout.

Another important note is that all the strategies are static,  i.e. except for the drop adherence and opponent modeling, they do not have a memory of the earlier moves, and they do not take information from the moves of the other players, leaving scope for further improvement. But for now, this provides us with a starting framework for simulation based analysis. Also, enabling drop adherence, atleast for random and defeat seeking strategies is not productive, as the game essentially ends very quickly due to their non-intelligent folds. Hence the main match-up between strategies are played without a drop adherence, and to estimate the drop adherence, we perform another 2 sets of match-ups with and without drop adherence, for the best performing strategy in the first match-up.

\section{Results}

To quantitatively assess the relative performance of different strategies, we conducted large-scale simulations of 1,000 games between every pair of strategies (each without drop-adherence). For each matchup, we recorded the winner, point gained, and several other performance metrics.

The following questions guide our empirical analysis.

\subsection*{Skill Analysis of strategies}
\begin{figure}[!htb]
    \centering
    \includegraphics[width=0.85\linewidth]{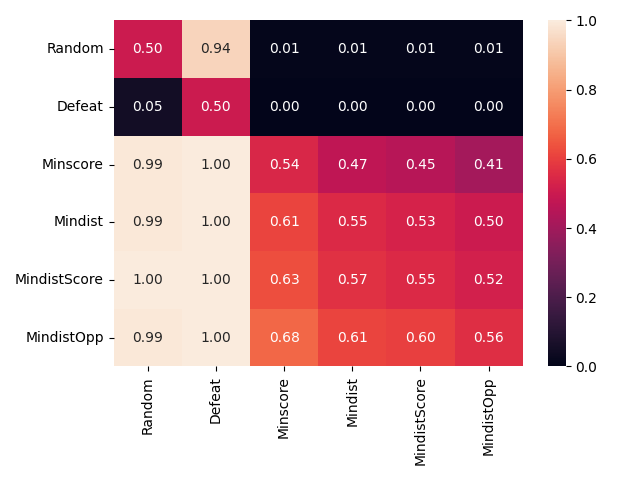}
    \caption{Pairwise Win Rate between strategies}
    \label{fig:winrate}
\end{figure}

We propose the win percentage as our primary metric for analysis. The \textit{win rate} is defined as the average proportion of wins of one strategy over 1,000 simulated games against another. Table~\ref{fig:winrate} summarizes the estimated win rates (for the first player, on the left) between all strategy pairs.

From Figure~\ref{fig:winrate}, we observe that \textbf{Mindist enhanced by Opponent modeling} consistently outperforms the others (both as a first player, signified by the last row being coordinate wise larger than all other rows, and as a second player, signified by last column smaller than every other column), achieving over 60\% win rate in most match-ups, suggesting a significant competitive advantage.

Another important point to notice is that Random and Defeat Seeking strategies also fulfills their goals (supporting the intuition), i.e. obtaining a win against Defeat Seeking and lose against others, and obtaining a loss against all, respectively. Hence an anti-optimal strategy (Defeat Seeking) may lose a game\footnote{In the game of Rummy, a player may never win by just not declaring, and in the current analysis, Defeat seeking is shown to win a few games despite having the chance to not declare. To clarify, here the defeat seeking strategy is acting naively, and if by the approach of defeat seeking described earlier, transforms the hand into a winning hand, defeat seeking may win a few matches against the random strategies (as it has more or less 70/80 rounds before a random strategy finishes with a valid meld}, and a skilled strategy (MinDistOpp) wins with positive edge, supporting skill element in the game of Rummy.

\subsection*{First mover's advantage}
Excluding the Random and Defeat Seeking strategies, we observe a strong 4-6\% edge for the first movers. To estimate the first mover's advantage, we assume first mover's advantage is different for different strategies (as observed also), and hence the win probability $p_{ij}$ between strategies $i$ and $j$, can be decomposed into:
$$p_{ij} = a_i+p^*_{ij}$$
where $i$ denotes the first player for the games considered, $a_i$ is the first mover's advantage for $i^{th}$ player, and $p^*_{ij}$ is the true skill component of the win probability.
Further, assuming $p^*_{ji} = 1-p^*_{ij}$ or $p^*_{ij} + p^*_{ji} = 1$, makes the problem estimable. Putting $j=i$, gives $\hat{a_i} = \frac{\hat{p_{ij}}+\hat{p_{ji}} -1}{2} = \hat{p_{ii}}-0.5$. From the data presented above, the estimates for $p^*_{ij}$s are given below. 

\begin{table}[H]
    \centering
    \begin{tabular}{p{1.75cm}p{1cm}p{1cm}p{1.5cm}p{1.5cm}}
        \toprule
        Strategies & $Minscore$ & $Mindist$ & $MindistScore$ & $MindistOpp$ \\
        \midrule
        $Minscore$ & 0.5 & 0.43 & 0.41 & 0.37 \\
        $Mindist$ & 0.55 & 0.5 & 0.48 & 0.45 \\
        $MindistScore$ & 0.58 & 0.52 & 0.5 & 0.47 \\
        $MindistOpp$ & 0.62 & 0.55 & 0.54 & 0.5 \\
        \bottomrule
    \end{tabular}
    \caption{True skill win probabilities\\ (first mover's advantage adjusted)}
    \label{tab:first.movers}
\end{table}

From the table above, the skill gradient is more prominent, showing $Minscore \prec Mindist \prec MindistScore \prec MindistOpp$. To test the significance of this additional skill components, we consider the following hypotheses:
$$H_0:p_{ij}=0.5 \quad vs \quad H_1:p_{ij}\neq 0.5$$\\
where $p_{ij}$ is the win probability of $i^{th}$ player against $j^{th}$ player. For $1000$ games, the standard error for the win rate (which is average of these trials) is $\sqrt{\frac{p(1-p)}{n}} = \sqrt{\frac{0.5\times0.5}{1000}} = 0.016$. Hence at a $95\%$ confidence level, under null hypotheses of of equal skill (i.e. $p_{ij}=0.5$), we get the confidence intervals as $(0.5 \pm 0.03) = (0.47,0.53)$. We observe all the adjusted win probabilities are outside this interval, implying \textbf{statistically significant skill gradient} at 0.05 level.

\subsection*{Gain analysis}
The \textit{gain} is defined as the point difference. 
Due to the presence of extreme values ($\pm$20 or $\pm$40 points), we adopt the median gain as a more robust measure of central tendency.

\subsection*{Drop adherence}
As the \textbf{MinDistOpp (Mindist enhanced by Opponent modeling)} is our best strategy, we have conducted another 1000 games between two MinDistOpp strategies, with and without drop adherence. The win percentage, mean and median gains are reported below.

\begin{table}[H]
    \centering
    \begin{tabular}{ccc}
        \toprule
        Strategies & $M\!DO\_drop$ & $M\!DO\_nodrop$ \\
        \midrule
        $M\!DO\_drop$ & - & 0.56, 4.63, 11.0 \\
        $M\!DO\_nodrop$ & 0.58, 3.49, 13.0 & - \\
        \bottomrule
    \end{tabular}
    \caption{MinDistOpp drop enabled vs drop disabled:\\Win percentage, mean and median gains resp.}
    \label{tab:drop.adherence}
\end{table}

As we can see, a strong first mover's advantage is observed here. To negate the first mover's advantage as before, we obtain a $7\%$ first movers advantage, and resulting true skill win probabilities are 0.49 and 0.51 respectively; from the confidence interval discussed before, we claim insignificant skill difference between drop enabled and drop disabled strategies. An important reason for this unintuitive result (drop adherence is verified as a skill by domain experts \cite{GAIM}) might be weak-drop adherence developed (drop hand if \emph{MinDist} for the hand is $>$ 4, hence rare cases). Dropping more frequently might increase the skill gap, but it then hampers the performance against other strategies (it can convert weaker hands against weaker strategies, drawing parallels from \href{https://www.chess.com/terms/odds-chess}{odds chess}). Hence this is a tuning parameter, which can be adjusted based on belief in opponent skill for optimal performance.

\subsection*{Analysis of Number of Rounds}
The number of rounds for a game can vary from 1 to 100, depending on strategies (even strategy pairs). It is expected if the random an defeat seeking strategies can not drop their hand, then the average number of rounds for them will be fairly close to 100, while for the skilled players the same number will be significantly smaller. Below we present the average number of moves between strategy pairs.

\begin{figure}[!htb]
    \centering
    \includegraphics[width=0.7\linewidth]{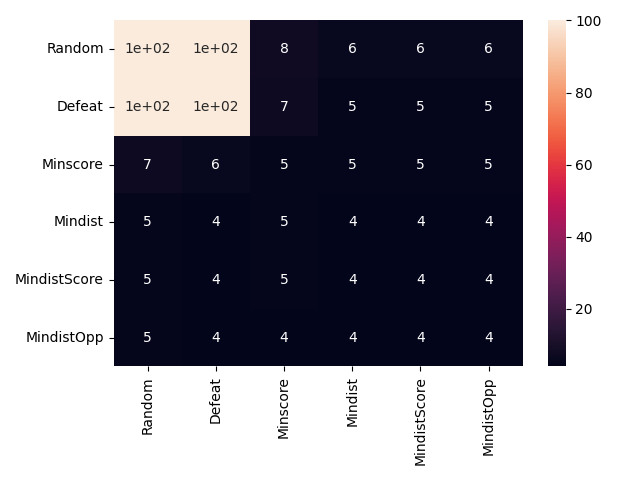}
    \caption{Mean number of rounds between strategy pairs}
    \label{fig:numrounds.table}
\end{figure}

We see that mean number of rounds is satisfying the aforementioned intuition. Also, the mean number of round for skilled players is significantly less than the upper bound, 100. Hence probability that the game will end next round significantly increases over each round, requiring an estimate for number of rounds required to win the game, hence promoting \emph{MinDist} as an hand evaluation metric.

\section{Discussion and Conclusion}
The simulation results clearly demonstrate a measurable skill hierarchy among the proposed strategies. The \textbf{MinDistOpp} strategy, which integrates hand optimization with opponent modeling, consistently achieves the highest win rates, confirming that adaptive reasoning and anticipatory play provide a decisive advantage in Rummy.

After adjusting for first-mover advantage, a distinct progression emerges: \textit{Minscore} $\prec$ \textit{Mindist} $\prec$ \textit{MindistScore} $\prec$ \textit{MindistOpp}. This gradient is statistically significant, underscoring that increasingly sophisticated decision rules translate directly into better performance.

The analysis of drop adherence, however, reveals no significant skill difference between the drop-enabled and drop-disabled variants of \textit{MinDistOpp}. This likely stems from a conservative drop rule; stronger adherence may improve outcomes but reduce robustness across diverse opponents, similar to odds-balancing in chess. Finally, the shorter game durations observed for skilled strategies indicate that higher competence not only improves success rates but also accelerates play. 

These results also point to clear avenues for refinement within the same framework. It is to note that \emph{MinScore} and \emph{MinDist} are based on the only the hands, and hence strategies built based on these metrics are static (i.e. only depends on that snapshot of hands, not on other information accumulated across a game, for example: the pile cards, opponent draws etc, which are only incorporated heuristically by hand modeling). The metrics can be enhanced to make them dynamic (or evolving) with respect to a game (i.e. incorporating discards and/or incorporate accumulative belief regarding opponent hand). To take an example, as \emph{MinDist} calculates the minimum possible alterations to make a given hand complete, the replaced cards can be any card (represented by \textit{super jokers}). But the cards seen on the pile stack are unavailable (atleast till the deck is not empty), and hence keeping this in mind, together with belief regarding other cards to be in opponent hands or in the available deck, the metric can be enhanced to capture the expected number of turns to make the hand complete, more accurately. Moreover, The clear separation between hand evaluation and strategic choice suggests that the proposed heuristics can serve as principled baselines for further adaptive or learning-based methods.

Overall, these results highlight that, despite inherent randomness, Rummy exhibits consistent and quantifiable skill components; supporting its treatment as a strategic, skill-based game amenable to formal analysis.

\bibliographystyle{IEEEtran}
\bibliography{rummy}

\end{document}